# Combining Minkowski and Chebyshev: New distance proposal and survey of distance metrics using k-nearest neighbours classifier

É.O. Rodrigues

*Department of Computer Science, Universidade Federal de Itajubá (UNIFEI), Minas Gerais - Brazil*

A R T I C L E   I N F O

A B S T R A C T

This work proposes a distance that combines Minkowski and Chebyshev distances and can be seen as an intermediary distance. This combination not only achieves efficient run times in neighbourhood iteration tasks in $\mathbb{Z}^2$, but also obtains good accuracies when coupled with the k-Nearest Neighbours (k-NN) classifier. The proposed distance is approximately 1.3 times faster than Manhattan distance and 329.5 times faster than Euclidean distance in discrete neighbourhood iterations. An accuracy analysis of the k-NN classifier using a total of 33 datasets from the UCI repository, 15 distances and values assigned to $k$ that vary from 1 to 200 is presented. In this experiment, the proposed distance obtained accuracies that were better than the average more often than its counterparts (in 26 cases out of 33), and also obtained the best accuracy more frequently (in 9 out of 33 cases).

## 1. Introduction

Distance metrics are widely used in data mining, visual computing, computer graphics, information analysis/retrieval, etc. From the mathematical point of view, distance is defined as a quantitative measurement of how far apart two objects are. Distance measurements that satisfy the metric properties are defined as metrics [2,9].

k-Nearest Neighbours (k-NN) is a lazy classifier that visits the k-nearest instances in the training dataset, computes the mode (the value that occurs most often), and uses this information to predict the label of unlabelled instances [4,17]. That is, a fruit that is surrounded by 9 apples and 1 grape is mostly probably an apple. k-NN relies on this premise, where the majority of the surrounding elements most probably indicates the class of the “unidentified object”, with an obvious margin of error that varies from problem to problem or dataset to dataset. The order of proximity of the elements is dictated by a distance measure.

k-NN is very simple and requires just two parameters. One of them is the distance measure that defines the nearest instances. The other parameter is the $k$ variable, which places an upper bound on how many nearest instances are visited. If k = 2, the labels of the 2 nearest instances are visited, where the majority of these labels indicates the prediction for the near unlabelled instance.

Some combinations of distance measures and $k$ values should work better than others. This is a known issue in classification, which is called parameter optimization. The best combination of input parameters varies according to the classification problem, and therefore it is very difficult to anticipate a good set of parameters without trying and analysing the obtained performance.

In this work, a survey and performance analysis of several distance metrics using k-NN in a total of 33 entirely numerical datasets of the UCI repository [14] is presented. Statistical analyses indicated that some distances outperform the remaining more often. Besides, the average value for $k$ across all the tested datasets is provided, which may assist in guessing a reasonable value for $k$ without relying on empirical analyses. Furthermore, a novel distance metric that combines Chebyshev and Minkowski distances is proposed, which produces an efficient classification performance in terms of accuracy and efficient means of iterating discrete neighbourhoods.

*E-mail address:* erickr@id.uff.br

This work is organized as follows. In the next section, a literature review is performed regarding distance metrics. Section 3 presents and discusses algorithms for neighbourhood iterations in 2 dimensional spaces using Chebyshev, Manhattan, Euclidean and the proposed distance. The run times achieved with each algorithm are compared in Table 1. Section 3.1 formally defines the proposed distance and proves that it is a metric for $p \geq 1$, $w_1 > 0$ and $w_2 > 0$. Section 3.2 illustrates how the proposal compares to other distances in two dimensions. Section 4 describes the obtained results. At last, Section 5 addresses the conclusions and discussions regarding future work.

## 2. Literature review

A metric space is a pair $(\chi, d)$ where $\chi$ is a set and $d$ is a mapping from $\chi \times \chi$ into $\mathbb{R}$, which satisfies the following conditions. Let $\chi$ be an arbitrary nonempty set and $d$ a real-valued function on the Cartesian product $\chi \times \chi$:

$$d : \chi \times \chi \rightarrow \mathbb{R} \tag{1}$$

and consider the following four properties, which hold for arbitrary points $x, y, z \in \chi$.

1. $d(x, y) \geq 0$ (symmetry)
2. $d(x, y) = 0$ iff $x = y$ (non-negativeness)
3. $d(x, y) = d(y, x)$ (positiveness)
4. $d(x, z) \leq d(x, y) + d(y, z)$ (triangle inequality)

A real-valued function $d$ on $\chi \times \chi$ that satisfies all four properties is a metric in $\chi$, and the properties themselves are called the metric axioms. A set $\chi$ equipped with a metric $d$ on $\chi \times \chi$ is a metric space, also denoted by the pair $(\chi, d)$. When $d$ satisfies all but the third property, it is called pseudometric. A distance function is any real-valued function $d$ on $\chi \times \chi$ that satisfies property 1 and 2. Similarly, when the fourth property is not satisfied, it is called semimetric [3].

**Example** Let $\chi$ be a non-empty set and $d$:

$$d(x, y) = \begin{cases} 1, & \text{if } x \neq y \\ 0, & \text{if } x = y \end{cases} \tag{2}$$

$(\chi, d)$ is an example of a discrete metric space.

The Euclidean distance [15] is the most commonly used metric when it comes to k-NN and other classification algorithms [16]. This may be associated to the fact that we are immersed in an Euclidean space, and therefore the Euclidean distance naturally comes to mind. However, other distances outperform the Euclidean distance in several occasions, which may be unintuitive at first glance [3].

Euclides once stated that the shortest distance between two points is a line. This assertion popularized the Pythagorean metric as Euclidean distance, although derived from the Pythagorean Theorem [2]. The Euclidean distance $d_2(x, y)$ between points $x$ and $y$, where $x, y \in \mathbb{R}^n$, is given by Eq. (3), where $n$ represents the number of dimensions.

$$d_2(x, y) = \sqrt{\sum_{i=1}^{n} (x_i - y_i)^2} \tag{3}$$

The Manhattan distance [5] was proposed by Hermann Minkowski in the late 19th century [2] and is defined as the sum of the absolute differences of the Cartesian coordinates, as shown in Eq. (4).

$$d_1(x, y) = \sum_{i=1}^{n} |x_i - y_i| \tag{4}$$

Later, Minkowski [1,6] included an exponent $p$ in its formulation. Minkowski distance, coined after him, is a generalization of both Euclidean ($p = 2$) and Manhattan ($p = 1$) distances. The metric conditions are satisfied as long as $p$ is equal or greater than 1. $p < 1$ violates the triangle inequality (fourth condition).

$$d_p(x, y) = \sqrt[p]{\sum_{i=1}^{n} |x_i - y_i|^p} \tag{5}$$

When $p$ reaches positive infinity, the Chebyshev distance is obtained, as shown in Eq. (6).

$$d_\infty(x, y) = \max_{i=1}^{n} |x_i - y_i| \tag{6}$$

These 4 distances metrics, Euclidean, Manhattan, Minkowski and Chebyshev, are the most popular metrics in research in a general basis.

The Squared Euclidean $d_{SD}$ is also known as the Sum of Squared Difference. This is the fundamental metric in least squares problems and linear algebra, and is shown in Eq. (7).

$$d_{SD}(x, y) = \sum_{i=1}^{n} (x_i - y_i)^2 \tag{7}$$

Finally, as a matter of comparison, the Canberra distance is a weighted version of the Manhattan distance initially proposed by Lance et al. [8] in the '60s, and is shown in Eq. (8).

$$d_{CAD}(x, y) = \sum_{i=1}^{n} \frac{|x_i - y_i|}{|x_i| + |y_i|} \tag{8}$$

### 2.1. Related works

Perlibakas [11] compares 14 distance measures in principal component analysis-based face recognition. Mahalanobis, among other three, were the best measures reported in the study. In a similar task, Kokare et al. [7] compare the performance of 9 measures for texture image retrieval, where the authors claim that Canberra and Bray–Curtis worked better than the remaining.

Lu et al. [10] varies the parameter $p$ of Minkowski distance and compares its performance in geographically weighted regressions. The authors indicate that, in fact, proper adjustments in $p$ leads to better accuracies.

Cha [2] published what is probably the most complete survey on similarity measures. The author defines that distance metrics and measures are similarity measures, while the opposite is not true. A total of 45 variations of similarity measures are addressed in his work. A general statistical analysis is performed, comparing the measures pairwise. No practical classification experiment is presented.

The literature still lacks an appropriate evaluation of distance metrics regarding a substantial number of datasets in classification tasks. This gap is covered by this work. Furthermore, along with this contribution, the main objective is proposing a novel distance metric, which is particularly efficient in discrete neighbourhood iterations.

## 3. Methodology

Iterating through the neighbourhood of a pixel can be done in a few different ways. If we consider the Chebyshev distance, the solution is straightforward. A loop iterates through the two lines above and below the central pixel and through the columns at its left and right. At each increment of the distance, the lines and columns are shifted by 1. Algorithm 1 illustrates the concept. After convergence, $N_d$ contains the pixels at distances $d$ from central pixel $(x_c, y_c)$.

It is slightly more difficult to perform neighbourhood iterations using the Manhattan distance. Manhattan is essentially the

**Algorithm 1:** Neighbourhood iteration using the Chebyshev distance.

**Data**: $I$ stands for an arbitrary image, $I(x, y)$ represents the pixel at position $(x, y)$. $D$ represents the maximal distance to be computed and $N_d$ is a set that contains all the pixels at distance $d$ from central pixel $(x_c, y_c)$.

1 $d \leftarrow 1$;
2 **while** $d < D$ **do**
3   **for** $(l \leftarrow -d; l \leq d; l \leftarrow l+1)$ **do**
4     $N_d \leftarrow N_d \cup I(x_c + l, y_c - d)$;
5     $N_d \leftarrow N_d \cup I(x_c + l, y_c + d)$;
6     **if** $l \neq d$ ***and*** $l \neq -d$ **then** `// Avoiding duplicates at corners`
7       $N_d \leftarrow N_d \cup I(x_c - d, y_c + l)$;
8       $N_d \leftarrow N_d \cup I(x_c + d, y_c + l)$;
9   $d \leftarrow d + 1$;

Chebyshev distance rotated by 45° in discrete 2D spaces. However, the implementation requires a few more operations and variables, which slightly increases the overall processing time. Algorithm 2 illustrates the process.

**Algorithm 2:** Neighbourhood iteration using the Manhattan distance.

**Data**: $I$ stands for an arbitrary image, $I(x, y)$ represents the pixel at position $(x, y)$. $D$ represents the maximal distance to be computed and $N_d$ is a set that contains all the pixels at distance $d$ from central pixel $(x_c, y_c)$.

1 $d \leftarrow 1$; $x_a \leftarrow 0$; $y_a \leftarrow 0$;
2 **while** $d < D$ **do**
3   **for** $(g \leftarrow 0; g < 4; g \leftarrow g+1)$ **do**
4     **switch** $g$ **do**
5       **case** *0* $x_a \leftarrow x_c$; $y_a \leftarrow y_c + d$;
6       **case** *1* $x_a \leftarrow x_c + d$; $y_a \leftarrow y_c$;
7       **case** *2* $x_a \leftarrow x_c$; $y_a \leftarrow y_c - d$;
8       **case** *3* $x_a \leftarrow x_c - d$; $y_a \leftarrow y_c$;
9     **for** $(l \leftarrow 0; l \leq d; l \leftarrow l+1)$ **do**
10       $N_d \leftarrow N_d \cup I(x_a, y_a)$;
11       **switch** $g$ **do**
12         **case** *0* $x_a \leftarrow x_a + 1$; $y_a \leftarrow y_a - 1$;
13         **case** *1* $x_a \leftarrow x_a - 1$; $y_a \leftarrow y_a - 1$;
14         **case** *2* $x_a \leftarrow x_a - 1$; $y_a \leftarrow y_a + 1$;
15         **case** *3* $x_a \leftarrow x_a + 1$; $y_a \leftarrow y_a + 1$;
16   $d \leftarrow d + 1$;

Any distance other than Chebyshev and Manhattan requires much more effort. In the Euclidean distance, a quantization/approximation is required. The distances can be pre-computed and stored in a kernel or processed in real time. Storing large amounts of data in kernels consume a significant amount of memory when computing large distances. Accessing this data also impacts negatively on the overall processing performance.

Algorithm 3 illustrates an arguably efficient way of iterating through the neighbourhood of the central pixel ($x_c$, $y_c$) using the Euclidean distance without storing the respective distances in a pre-built kernel. Two early breaks are employed to speed up the computation.

In average, Chebyshev distance (Algorithm 1) iterates a neighbourhood until $D = 2500$ in 0.098 s. Manhattan (Algorithm 2), on the other hand, takes approximately 0.146 s to iterate through the

**Algorithm 3:** Neighbourhood iteration using the Euclidean distance.

**Data**: $I$ stands for an arbitrary image, $I(x, y)$ represents the pixel at position $(x, y)$. $D$ represents the maximal distance to be computed and $N_d$ is a set that contains all the pixels at distance $d$ from central pixel $(x_c, y_c)$. $\delta$ must be replaced by $abs(\sqrt{(x_c - x_a)^2 + (y_c - y_a)^2}) = d$, where *abs* represents the absolute integer value.

1 $d \leftarrow 1$; $x_a \leftarrow 0$; $y_a \leftarrow 0$;
2 **while** $d < D$ **do**
3   **for** $(ls \leftarrow 0; ls < d/2; ls \leftarrow ls+1)$ **do**
4     foundFirst $\leftarrow$ false;
5     **for** $(l \leftarrow d; l \geq 0; l \leftarrow l-1)$ **do**
6       finished $\leftarrow$ false;
7       **for** $(ln \leftarrow l; \ln \geq -l; \ln \leftarrow \ln - (2 * l))$ **do**
8         $x_a \leftarrow x_c + \ln$; $y_a \leftarrow y_c - d + ls$;
9         **if** $\delta$ **then**
10           $N_d \leftarrow N_d \cup I(x_a, y_a)$;
11           foundFirst $\leftarrow$ true;
12         **else** finished $\leftarrow$ foundFirst;
13         $y_a \leftarrow y_c + d - ls$;
14         **if** $\delta$ **then** $N_d \leftarrow N_d \cup I(x_a, y_a)$;
15         **if** $ln \neq d$ **then** `// duplicates`
16           $x_a \leftarrow x_c - d + ls$; $y_a \leftarrow y_c + \ln$;
17           **if** $\delta$ **then** $N_d \leftarrow N_d \cup I(x_a, y_a)$;
18           $x_a \leftarrow x_c + d - ls$;
19           **if** $\delta$ **then** $N_d \leftarrow N_d \cup I(x_a, y_a)$;
20         **if** *ln = 0* **then** break;
21       **if** *finished* **then** break;
22   $d \leftarrow d + 1$;

same neighbourhood. Euclidean (Algorithm 3), at last, requires a total of 36.125 s, considering the same task.

When it comes to optimization, especially in Graphics Processing Unit (GPU) computing, these differences on the performance substantially impact overall processing times. Besides, Manhattan does not add any information to the neighbourhood iteration in regard to Chebyshev. As previously stated, Manhattan is the plain Chebyshev rotated by 45° in 2D discrete spaces.

On the other hand, using the Euclidean distance implies a huge performance burden, where its implementation is approximately 368 times slower than Chebyshev for $D = 2500$, and it gets even worse as $D$ increases. This work proposes an intermediate distance between Chebyshev and Euclidean that adds information, as opposed to Manhattan, and is far more efficient when it comes to neighbourhood iterations.

This distance was accidentally identified while attempting to improve a previous work [13]. This previous work perform iterations through the neighbourhood of images using GPUs, which heavily rely on low level operations. As previously stated, something closer to the Euclidean distance was desired, but the reported time burden was still incompatible with practical cases. Algorithm 4 shows the iteration process of the proposed distance.

The proposed distance almost achieves Chebyshev-like time performances and provides an order of iteration that is similar to the one obtained with the Euclidean distance. The result of the proposed distance in $\mathbb{R}^2$ for $p = 1$ is an octagon (Chebyshev and Manhattan are squares). That is, the octagon relates to circles more than squares, obtained with Manhattan and Chebyshev distances. Table 1 shows the processing times obtained with each one of the algorithms (Algorithms 1–4) as the size of the neighbourhood in-

**Algorithm 4:** Neighbourhood iteration using the proposed distance ($w_1 = w_2 = p = 1$).

**Data**: $I$ stands for an arbitrary image, $I(x, y)$ represents the pixel at position $(x, y)$. $D$ represents the maximal distance to be computed and $N_d$ is a set that contains all the pixels at distance $d$ from central pixel $(x_c, y_c)$.

1 $d \leftarrow 1$; $d_a \leftarrow d$;
2 **while** $d < D$ **do**
3 $\quad N_{d_a} \leftarrow N_{d_a} \cup I(x_c + d, y_c)$;
4 $\quad N_{d_a} \leftarrow N_{d_a} \cup I(x_c - d, y_c)$;
5 $\quad N_{d_a} \leftarrow N_{d_a} \cup I(x_c, y_c + d)$;
6 $\quad N_{d_a} \leftarrow N_{d_a} \cup I(x_c, y_c - d)$;
7 $\quad$ **for** $(l \leftarrow 1;\ l \leq d;\ l \leftarrow l + 1)$ **do**
8 $\quad\quad N_{(d_a+l)} \leftarrow N_{(d_a+l)} \cup I(x_c + d, y_c - l)$;
9 $\quad\quad N_{(d_a+l)} \leftarrow N_{(d_a+l)} \cup I(x_c + d, y_c + l)$;
10 $\quad\quad N_{(d_a+l)} \leftarrow N_{(d_a+l)} \cup I(x_c - d, y_c - l)$;
11 $\quad\quad N_{(d_a+l)} \leftarrow N_{(d_a+l)} \cup I(x_c - d, y_c + l)$;
12 $\quad\quad$ **if** $l \neq d$ **then**
13 $\quad\quad\quad N_{(d_a+l)} \leftarrow N_{(d_a+l)} \cup I(x_c - l, y_c + d)$;
14 $\quad\quad\quad N_{(d_a+l)} \leftarrow N_{(d_a+l)} \cup I(x_c + l, y_c + d)$;
15 $\quad\quad\quad N_{(d_a+l)} \leftarrow N_{(d_a+l)} \cup I(x_c - l, y_c - d)$;
16 $\quad\quad\quad N_{(d_a+l)} \leftarrow N_{(d_a+l)} \cup I(x_c + l, y_c - d)$;
17 $\quad d_a \leftarrow d_a + d + 1$;
18 $\quad d \leftarrow d + 1$;

**Table 1**
Processing times (s) for each distance (Algorithms 1–4).

| Sizes (D) | Chebyshev | Manhattan | Euclidean | Rodrigues |
|---|---|---|---|---|
| 500 | 0.0022 | 0.0024 | 0.3069 | 0.0022 |
| 1000 | 0.0109 | 0.0116 | 2.3290 | 0.0115 |
| 1500 | 0.0295 | 0.0407 | 7.8285 | 0.0318 |
| 2000 | 0.0583 | 0.0829 | 18.5112 | 0.0624 |
| 2500 | 0.0985 | 0.1460 | 36.1253 | 0.1050 |
| 3000 | 0.1517 | 0.2228 | 62.2829 | 0.1599 |
| 3500 | 0.2377 | 0.3397 | 98.8083 | 0.2454 |
| 4000 | 0.3031 | 0.4672 | 147.6250 | 0.3166 |
| 4500 | 0.4030 | 0.5923 | 204.2881 | 0.4243 |

creases. These results were obtained using Java and an Intel i7-7700HQ clocked at 2.8 GHz under the same conditions, averaged over 100 runs.

Table 1 evidences that the proposed distance is in fact faster than Manhattan and Euclidean distances. In average, it was 1.3 times faster than Manhattan and 329.5 times faster than Euclidean. In the following sub-section, a formal definition of this distance is presented.

### 3.1. Distance Definition

The proposed distance is a combination of Chebyshev and Minkowski distances, weighted by $w_1$ and $w_2$, as shown in Eq. (9). As $w_1$ increases in regard to $w_2$, the distance becomes more like Minkowski. Contrariwise, it converges towards Chebyshev. As a remark, it is assumed that $w_1, w_2, p \in \mathbb{R}$. Algorithm 4 considers $w_1 = w_2 = p = 1$.

$$d_{w_1,w_2,p}(x, y) = w_1\, d_p(x, y) + w_2\, d_\infty(x, y)$$
$$\vdots \tag{9}$$
$$d_{w_1,w_2,p}(x, y) = w_1 \sqrt[p]{\textstyle\sum_{i=1}^{n} |x_i - y_i|^p} + w_2 \max_{i=1}^{n} |x_i - y_i|$$

As two distance metrics are being summed up, it is straightforward to infer from the formulation that as long as $w_1, w_2 > 0$ and $p \geq 1$, all the distance metric conditions are satisfied, including the triangle inequality.

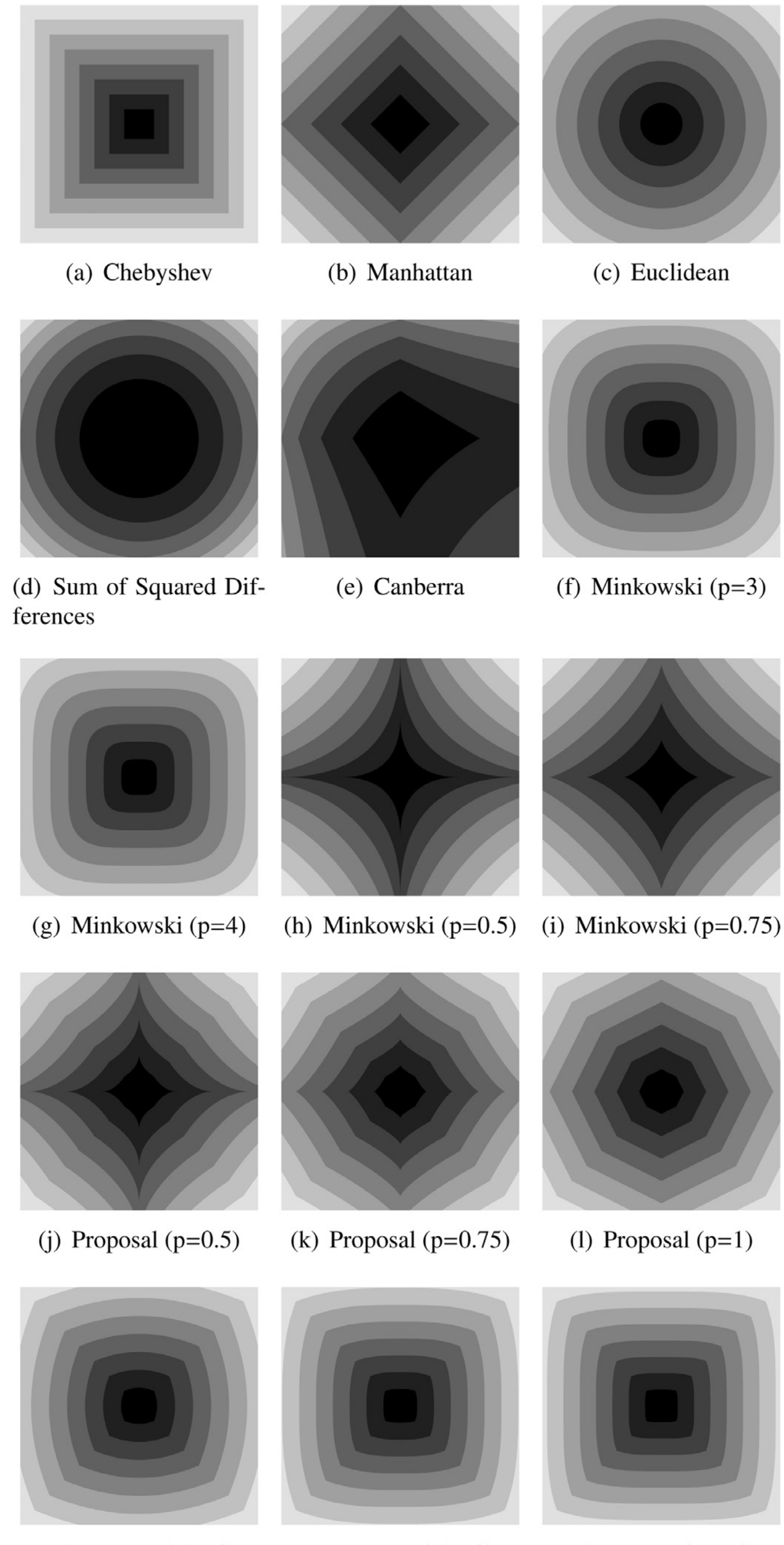

**Fig. 1.** Distances in $\mathbb{R}^2$. The distances are computed from the central element in the image. Light shades of gray indicate greater distances. In the cases where $p < 1$, the respective distances are not metrics. In all images, $w_1 = w_2 = 1$ is respected.

In what follows, Proof 3.1 demonstrates in detail how the metrics conditions are satisfied for $d_{w_1,w_2,p}(x, y)$.

**Proof 3.1.** $d_{w_1,w_2,p}(x, y)$ is a metric as long as $p \geq 1$, $w_1 > 0$ and $w_2 > 0$ are satisfied.

① $d_{w_1,w_2,p}(x, y) \geq 0 \rightarrow d_p(x,\ y)$ and $d_\infty(x,\ y)$ are metrics themselves and therefore provide values equal or greater than 0, their sum is also equal or greater than 0.

**Table 2**
Obtained results (mean of 33 numerical datasets from the UCI repository). For all cases, $w_1 = w_2 = 1$.

| Distances | Mean Acc | Mean TP | Mean TN | Mean k | Max k | *P*-Value | Better than Average | Best |
|---|---|---|---|---|---|---|---|---|
| Euclidean Distance | 0.823 | 0.768 | 0.867 | 11.2 | 49 | 0.552 | 19 | 6 |
| Chebyshev Distance | 0.786 | 0.733 | 0.860 | 14.4 | 108 | **0.129** | 9 | 4 |
| Manhattan Distance | 0.834 | 0.780 | 0.878 | 10.7 | 70 | 0.787 | 24 | 4 |
| Minkowski Distance ($p = 0.5$) | **0.842** | **0.789** | **0.884** | 12.3 | 97 | 1.000 | 25 | **9** |
| Minkowski Distance ($p = 0.75$) | 0.837 | 0.785 | 0.882 | 10.8 | 68 | 0.874 | 24 | 4 |
| Minkowski Distance ($p = 3$) | 0.821 | 0.766 | 0.867 | 10.2 | 47 | 0.516 | 16 | 6 |
| Minkowski Distance ($p = 4$) | 0.820 | 0.766 | 0.867 | 10.5 | 47 | 0.490 | 18 | 8 |
| Canberra Distance | 0.695 | 0.627 | 0.828 | 24.1 | 190 | **0.004** | 10 | 8 |
| Sum of Squared Difference | 0.823 | 0.768 | 0.867 | 11.2 | 49 | 0.552 | 19 | 6 |
| Proposed Rodrigues Distance ($p = 0.5$) | 0.840 | 0.788 | **0.884** | 10.8 | 68 | 0.950 | **26** | 5 |
| Proposed Rodrigues Distance ($p = 0.75$) | 0.835 | 0.781 | 0.879 | 10.8 | 66 | 0.816 | **26** | 3 |
| Proposed Rodrigues Distance ($p = 1$) | 0.829 | 0.776 | 0.875 | 11.7 | 101 | 0.677 | 23 | **9** |
| Proposed Rodrigues Distance ($p = 2$) | 0.821 | 0.766 | 0.867 | 11.4 | 45 | 0.519 | 17 | 4 |
| Proposed Rodrigues Distance ($p = 3$) | 0.820 | 0.768 | 0.868 | 10.9 | 47 | 0.501 | 18 | 5 |
| Proposed Rodrigues Distance ($p = 4$) | 0.820 | 0.765 | 0.867 | 10.8 | 47 | 0.489 | 16 | 8 |

② $d_{w_1,w_2,p}(x, y) = 0 \Leftrightarrow x = y \rightarrow$ $d_p(x, y)$ and $d_\infty(x, y)$ both return 0 if and only if $x = y$. The sum of these two results in 0 only if both are 0.

③ $d_{w_1,w_2,p}(x, y) = d_{w_1,w_2,p}(y, x) \rightarrow 0$, It is known *a priori* that both metrics $d_p(x, y)$ and $d_\infty(x, y)$ satisfy ③. Therefore, $d_p(x, y) = d_p(y, x)$ and $d_\infty(x, y) = d_\infty(y, x)$, which proves that $w_1\, d_p(x, y) + w_2\, d_\infty(x, y) = w_1\, d_p(y, x) + w_2\, d_\infty(y, x)$.

④ $d_{w_1,w_2,p}(x, z) \leq d_{w_1,w_2,p}(x, y) + d_{w_1,w_2,p}(y, z) \rightarrow$ Once again, as both inequalities $d_p(x, z) \leq d_p(x, y) + d_p(y, z)$ and $d_\infty(x, z) \leq d_\infty(x, y) + d_\infty(y, z)$ are true, consequently, $w_1\, d_p(x, z) + w_2\, d_\infty(x, z) \leq w_1\, d_p(x, y) + w_2\, d_\infty(x, y) + w_1\, d_p(y, z) + w_2\, d_\infty(y, z)$ is also satisfied. □

### 3.2. Graphical Analysis in $\mathbb{R}^2$

Fig. 1 shows how the proposed distance compares to other distances in $\mathbb{R}^2$. The proposed distance is in fact a mixture of Chebyshev and Manhattan, as shown in Fig. 1-(l). It is possible to argue that the distance is an intermediate step between Manhattan/Chebyshev and Euclidean, and therefore it adds information. Furthermore, it requires far less processing power (Algorithm 4) in regard to Euclidean or even Manhattan, as shown in the beginning of Section 3.

## 4. Experimental results

In this section, the performance of the previously addressed distances is evaluated. Experiments were performed using the k-Nearest Neighbours algorithm. All the 33 numerical datasets (the ones that do not contain any categorical attributes) from the UCI repository [14] were employed in the experiments.

Categorical attributes are widely used with k-NN and other classifiers that are supposedly made to work with numerical values. The commonly employed assumption is that if the categorical value is equal, the distance between these two instances is said to be 1. Otherwise, the categorical distance is 0. Although this assumption works well and enables the processing of categorical datasets by numerical classifiers, it could bias the experiments. If an arbitrary dataset contains 9 categorical and 1 numerical attributes, the distance measurement would be applied in just one of the attributes (1 dimensional). These assumptions could improve the performance of a certain distance in detriment of others by chance more frequently. To avoid biasing the overall results, categorical datasets were disregarded from the experiments.

The first three columns of Table 2 show the mean accuracy, true positive and true negative rates achieved by each distance. Accuracy is defined as the number of true positives and true negatives divided by the total population. The last five columns concern to the Mean *k*, Max *k*, *P*-value, how many times the accuracy obtained with the distance was better than the average accuracy, and the amount of occasions where the accuracy obtained with the respective distance was the best one (equal to the maximal obtained accuracy).

Values from 1 to 200 were assigned to the *k* variable for each combination of distance and dataset. The *k* value that achieved the best accuracy was selected. The fifth column (Mean k) represents the average *k* chosen over all the 33 datasets for each one of the distances. The sixth column (Max *k*) represents the maximal *k* that was chosen with the respective distance, also over the 33 occasions.

In the seventh column, *P*-value is computed considering the accuracy values of each distance in relation to the accuracy obtained with Minkowski distance where p = 0.5, which was the distance that obtained the best mean accuracy (highlighted in bold in the second column of Table 2).

Surprisingly, $p = 0.5$ achieved the best mean accuracy with Minkowski and the proposed distance, closely followed by $p = 0.75$ (although both of them are not metrics). True positive and true negative rates should not be compared separately, either accuracy or both of them should be considered for a fair comparison.

The mean *k* varied over distances but stayed within the 10-11 margin in most cases. Max *k* and mean *k* were substantially larger with Canberra distance, which is less sensitive near the origin.

*P*-values indicate that Canberra was significantly worse than the distance that obtained the highest mean accuracy (Minkowski $p = 0.5$), where its *P*-value was less than 0.005. The *P*-value of Chebyshev is not as weak as Canberra, but also indicates that Chebyshev is not as competitive in general as the remaining distances, which can also be stated by its mean accuracy.

At last, the proposed distance obtained accuracies that were better than the average more frequently (in 26 out of 33 cases) and, along with Minkowski ($p = 0.5$), more often obtained accuracies that were the best (in 9 out of 33 cases). In contrast to Minkowski, the proposed distance was more often the best with $p = 1$, which is also the value used for *p* in Algorithm 4, proving that it is in fact a powerful distance metric in terms of time efficiency and accuracy.

## 5. Conclusions

This work proposes and highlights the fact that combining Minkowski and Chebyshev distances results in an efficient distance metric in terms of processing times and accuracy. The idea first originated from an initiative of iterating through the neighbourhood of images in low level instruction environments such as GPUs [13]. The proposed distance proved to be approximately 1.3 times faster than Manhattan and 329.5 times faster than Euclidean, while also being more similar to Euclidean in contrast to Manhattan and Chebyshev.

An extensive experiment regarding 33 datasets of the UCI repository was conducted in order to evaluate the efficiency of the proposed distance using k-NN. This novel distance achieved accuracies that were better than the average more frequently than its counterparts (26 cases in 33). Furthermore, it obtained the best accuracy more often (9 cases in 33).

As a remark, a paired t-test indicated that Canberra is significantly worse than the remaining distances in average. Chebyshev also achieved a low P-value in regard to the remaining distances that, coupled to its mean accuracy, indicate that Chebyshev may in fact be worse in average in terms of accuracy than all distances but Canberra. The experiments also indicate that the average $k$ for the k-NN algorithm lays near 12, averaged across all distances.

The proposed distance enables a wealth of possibilities for applications in data mining and visual computing. Future works may focus on evaluating this proposal with other classifiers that rely on distances or neighbourhood iterations such as the ones based on mathematical morphology. Finally, all the datasets and source code used in this work are publicly available at [12].

## References

[1] A.R. Ali, M.S. Couceiro, A.E. Hassanien, D.J. Hemanth, Fuzzy c-means based on minkowski distance for liver ct image segmentation, Intell. Decis. Technol. 10 (4) (2016) 393–406.
[2] S. Cha, Comprehensive survey on distance/similarity measures between probability density functions, Int. J. Math. Models Methods Appl. Sci. (2007) 300–307.
[3] A. Conci, C.S. Kubrusly, Distance between sets: a survey, Adv. Math. Sci. Appl. 17 (2017) 1343–4373.
[4] N. Garcia-Pedrajas, J.A.R. Castillo, G. Cerruela-Garcia, A proposal for local k values for k -nearest neighbor rule, IEEE Trans. Neural Networks Learn. Syst. 28 (2) (2017) 470–475.
[5] L. Greche, M. Jazoui, N. Es-Sbai, A. Majda, A. Zarghili, Comparison between euclidean and manhattan distance measure for facial expressions classification, Wireless Technol., Embedded Intell. Syst. (WITS) (2017), doi:10.1109/WITS.2017.7934618.
[6] R. Kamimura, O. Uchida, Greedy network-growing by minkowski distance functions, in: Proceedings of the 2004 IEEE International Joint Conference on Neural Networks, 2004.
[7] M. Kokare, B.N. Chatterji, P.K. Biswas, Comparison of similarity metrics for texture image retrieval, Conf. Converg. Technol. Asia-Pacific Region (2003), doi:10.1109/TENCON.2003.1273228.
[8] G.N. Lance, W.T. Williams, Computer programs for hierarchical polythetic classification (similarity analyses), Comput. J. 9 (1) (1966) 60–64, doi:10.1093/comjnl/9.1.60.
[9] F.W. Lawvere, Metric spaces, generalized logic, and closed categories, Reprints Theory Appl. Categories 1 (2002) 1–37.
[10] B. Lu, M. Charlton, C. Brunsdon, P. Harris, The minkowski approach for choosing the distance metric in geographically weighted regression, Int. J. Geograph. Inf. Sci. 30 (2) (2015) 351–368.
[11] V. Perlibakas, Distance measures for pca-based face recognition, Pattern Recognit. Lett. 25 (6) (2004) 711–724.
[12] Rodrigues, E. O., 2017. https://github.com/Oyatsumi/RodriguesDistance
[13] E.O. Rodrigues, L. Torok, P. Liatsis, J. Viterbo, A. Conci, k-ms: a novel clustering algorithm based on morphological reconstruction, Pattern Recognit. 66 (2017) 392–403.
[14] UCI, 2017. Index of /datasets/uci/arff/.
[15] L. Wang, Y. Zhang, J. Feng, On the Euclidean distance of images, IEEE Trans. Pattern Anal. Mach. Intell. 27 (8) (2005) 1334–1339.
[16] K.Q. Weinberger, L.K. Saul, Distance metric learning for large margin nearest neighbor classification, J. Mach. Learn. Res. 10 (2009) 207–244.
[17] I.H. Witten, E. Frank, M.A. Hall, C.J. Pal, Data Mining: Practical Machine Learning Tools and Techniques, Morgan Kaufmann, 2016.